\documentclass{article}
\usepackage{iclr2027_conference,times}
\usepackage{amsmath,amssymb}
\usepackage{booktabs,tabularx}
\usepackage{graphicx}
\usepackage{hyperref}
\usepackage{url}
\usepackage{algorithm}
\usepackage{algpseudocode}
\usepackage[table]{xcolor}

\algblockdefx{Parallel}{EndParallel}
  {\textbf{parallel branches}}
  {\textbf{end parallel}}

\newcommand{\method}{DeltaTTT}

\newcommand{\silu}{\operatorname{SiLU}}

\title{DeltaTTT: Layerwise Optimization\\for Nonlinear Recurrent Memory}
\author{
Yining Li$^{1,2}$ \quad
Dongchen Han$^{2}$ \quad
Jie Fu$^{3}$ \quad
Gao Huang$^{2,\dagger}$ \\[0.5em]
{\normalfont
$^{1}$Qiuzhen College, Tsinghua University \quad
$^{2}$LeapLab, Tsinghua University} \\
{\normalfont $^{3}$IQuest Research}
}
\iclrfinalcopy

\begin{document}
\maketitle
\fancyhead{}
\lhead{Preprint}

\begingroup
\renewcommand{\thefootnote}{\ensuremath{\dagger}}
\footnotetext{Corresponding author.}
\endgroup

\begin{abstract}
Sequential test-time training adapts a memory network through successive updates, each computing an inner-loop gradient based on the network's previous state. Intuitively, this state dependence should allow each update to account for what the memory has already learned and better incorporate new information. However, we find that this expected advantage does not consistently materialize in nonlinear memories: a fixed-base parallel TTT baseline outperforms its serial counterpart. Our exploratory experiments point to a key underlying difficulty: nonlinear memories can be harder to optimize than linear ones within a single pass over the sequence. To alleviate this optimization difficulty, we introduce \method{}, which replaces joint inner-loop optimization of a two-layer memory network with layerwise learning. Each layer is assigned a local prediction target and updated through a state-dependent delta rule. This formulation retains a nonlinear readout while enabling chunkwise parallel computation. Experiments on DeltaNet and LaCT backbones show improvements in language modeling and retrieval over their recurrent baselines.

\end{abstract}

\section{Introduction}
\label{sec:introduction}


In sequence modeling, Test-Time Training (TTT) compresses past
key--value pairs into the parameters of a small
inner network through online updates \citep{sun2024ttt}.
By default, the updates are sequential, each computing an
inner-loop gradient based on the network's previous state.
Intuitively, this state dependence should improve memory
learning: each update can account for what the network has
already learned and better incorporate new information.


This intuition is supported by results on linear memories.
DeltaNet replaces the state-independent additive writes of
linear attention with residual corrections based on the
current memory, achieving stronger associative
recall and supporting the value of state-dependent updates
\citep{yang2024deltanet}.
However, this expected advantage does not consistently
materialize for nonlinear memories.
In a LaCT-derived language model \citep{zhang2025lact}, we
compare serial updates, which evaluate each chunk at the
current fast weights, with a fixed-base parallel baseline,
which evaluates updates at a shared initial state and
accumulates them causally.
We find that the parallel baseline achieves better downstream results despite computing its updates without
considering earlier writes.
This result motivates us to revisit how nonlinear memory is optimized in serial TTT.

To examine memory optimization more directly, we fit standalone
linear and two-layer MLP memories to fixed K/V sequences extracted
from a pretrained full-attention model, using key-value reconstruction
as the objective. A linear memory can achieve lower reconstruction
error within a single epoch, while the MLP's advantage emerges with
additional epochs. These observations suggest that greater
expressivity alone is insufficient: the write rule must also
realize that capacity within the limited inner-update budget, usually one epoch for TTT.

We therefore ask: if jointly optimizing a nonlinear memory can
be less effective than optimizing a linear one within a single
pass, can we optimize it more effectively layer by layer? This motivates DeltaTTT, which retains a nonlinear multilayer memory while equipping each linear layer with a local reconstruction target and its own delta-rule update. Given its inputs and targets, each layer admits efficient chunkwise parallel evaluation, while the nonlinear composition and dependencies between layers are preserved. This design aims to combine the expressive capacity of nonlinear memories with effective single-pass optimization and the computational efficiency of the delta rule.


Our contributions are threefold.
1) We show that serial nonlinear TTT can underperform
fixed-base parallel TTT, and use fixed-K/V reconstruction
experiments to highlight the difficulty of optimizing
nonlinear memories within a single pass.
2) To alleviate this optimization difficulty, we propose \method{}, which replaces joint inner-loop
optimization of a two-layer memory network with layerwise
learning, assigning each layer a local prediction target
and a state-dependent delta-rule update.
This formulation retains a nonlinear readout and enables
chunkwise parallel computation while preserving exact
tokenwise updates.
3) We evaluate \method{} with 0.5B-parameter language models
trained on 92.5B tokens, demonstrating improvements in
language modeling and retrieval over recurrent baselines
on both DeltaNet and LaCT backbones.
\begin{figure}
    \centering
    \includegraphics[width=\linewidth]{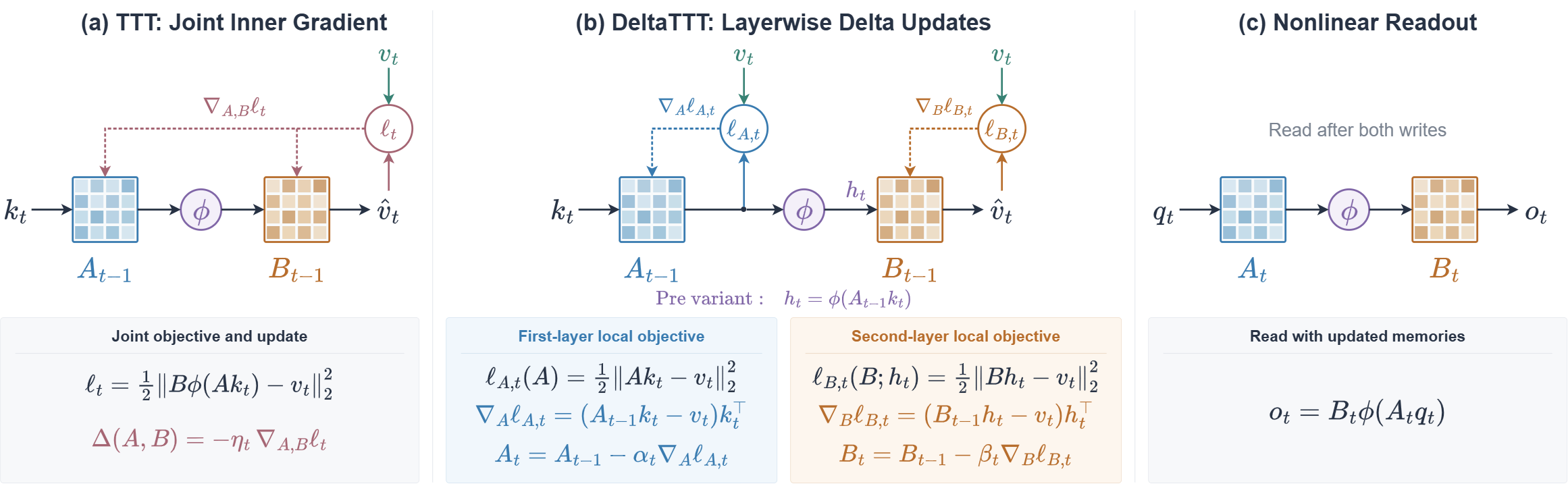}
    \caption{\textbf{Joint and layerwise optimization of nonlinear recurrent memory.}
(a) TTT jointly optimizes both layers through a shared reconstruction objective.
(b) DeltaTTT assigns a local reconstruction objective to each layer and performs layerwise delta-rule updates.
(c) Both methods use the nonlinear readout after the writes.}
    \label{fig:DeltaTTT}
\end{figure}

\section{Background}
\label{sec:background}

\noindent\textbf{Test-Time Training.}
Test-Time Training (TTT) represents sequence context through
the parameters $W$ of an inner network $f_W$, typically an
MLP\citep{sun2024ttt,han2026vit3}.
Given query, key, and value features $(Q,K,V)$, TTT compresses
key--value associations into $W$ through inner-loop optimization:
\begin{equation}
W \leftarrow W-\eta\nabla_W\mathcal{L}\bigl(f_W(K),V\bigr),
\end{equation}
where $\mathcal{L}$ is the inner learning objective and
$\eta$ is the learning rate.
Queries retrieve information from the resulting memory through
\begin{equation}
O=f_W(Q).
\end{equation}
As the sequence is processed, these updates allow the inner
network to accumulate context in a fixed-size parameter state.
With a fixed-size inner network and a single pass over the
sequence, the total read and update cost scales linearly
with sequence length.

\paragraph{Serial and fixed-base parallel TTT.}
Our serial baseline follows the chunkwise TTT formulation
of LaCT \citep{zhang2025lact}.
Let $\mathcal{L}_c(W)$ denote the inner loss on chunk $c$
evaluated at memory parameters $W$.
We maintain an unnormalized accumulator $S_c$ and a normalized
memory $W_c$, initialized with $S_0=W_0$.
For clarity, we omit momentum.
Serial TTT evaluates each gradient at the memory updated
by the preceding chunks, whereas fixed-base parallel TTT
evaluates every gradient at the shared initialization:
\begin{align}
\text{Serial:}\qquad
S_c
&=
S_{c-1}-\eta\nabla\mathcal{L}_c(W_{c-1}),
\qquad W_c=\mathcal{N}(S_c),
\label{eq:serial}\\
\text{Fixed-base:}\qquad
S_c
&=
S_{c-1}-\eta\nabla\mathcal{L}_c(W_0),
\qquad W_c=\mathcal{N}(S_c),
\label{eq:parallel}
\end{align}
where $\eta$ is the inner learning rate and $\mathcal{N}$
means normalization.
Serial gradients depend on preceding updates, while fixed-base
gradients can be computed independently across chunks and
accumulated through a causal prefix sum.
Both baselines read before writing the current chunk,
using $o_t=f_{W_{c-1}}(q_t)$ for tokens $t$ in chunk $c$.

\paragraph{The linear case: DeltaNet and Linear Attention.}
The same framework recovers DeltaNet by using a linear memory
$f_W(k)=Wk$, the squared loss
$\ell(y,v)=\frac{1}{2}\|y-v\|_2^2$, and one token per update
(chunk size=1).
Indexing states by token, serial gradient descent gives
\begin{equation}
\begin{aligned}
W_t
&=W_{t-1}-\eta_t\nabla\mathcal{L}_t(W_{t-1})\\
&=W_{t-1}
+\eta_t(v_t-W_{t-1}k_t)k_t^\top.
\end{aligned}
\label{eq:delta}
\end{equation}
This is precisely the delta-rule update used in DeltaNet.
For supplied keys, values, and learning rates, this recurrence
is affine in the preceding state, enabling efficient
chunkwise parallel evaluation \citep{yang2024deltanet}.
DeltaNet thus combines tokenwise update granularity
(chunk size=1) with computational parallelism, preserving
the exact sequence of state-dependent updates.
Applying fixed-base evaluation to the same objective instead
yields
\begin{equation}
{W}_t
=
W_0+\sum_{i=1}^{t}
\eta_i(v_i-W_0k_i)k_i^\top.
\label{eq:fixed_base_linear}
\end{equation}
With $W_0=0$ and $\eta_i=1$, this reduces to
${W}_t=\sum_{i=1}^{t}v_i k_i^\top$,
which is exactly linear attention.
Thus, DeltaNet and additive linear attention correspond to
serial and fixed-base parallel TTT respectively within linear memory. 

\section{Motivation: Effective Optimization in a Single Pass}
\label{sec:motivation}

Serial TTT is intuitively promising because this state dependence should allow each update to account for what the memory has already learned and better incorporate new information. This expectation is supported by DeltaNet, which uses the residual $v_t-W_{t-1}k_t$ to perform state-dependent correction (Equation~\ref{eq:delta}) and achieves stronger associative recall than additive linear attention \citep{yang2024deltanet}. Surprisingly, we observe the opposite in nonlinear TTT: serial updates achieve a single-needle accuracy of 28.25, compared with 37.87 for fixed-base parallel updates (Table~\ref{tab:motivation}). We consider two possible explanations for this discrepancy.

\begin{table}[t]
\caption{Serial and parallel TTT with SwishGLU inner models,
0.5B parameters, and 92.5B training tokens.
Single denotes mean single-needle retrieval accuracy across
S1/S2/S3;
32k denotes the mean accuracy across S1/S2/S3 at 32k context length.
}
\label{tab:motivation}
\centering
\begin{tabular}{lrrr}
\toprule
Update rule & Loss $\downarrow$ & Single $\uparrow$ & 32k $\uparrow$ \\
\midrule
Serial  & 1.99 & 28.25 & 11.13 \\
Parallel & 1.98 & 37.87 & 30.87 \\
\bottomrule
\end{tabular}
\end{table}

\begin{figure}[t]
\centering
\includegraphics[width=0.9\linewidth]{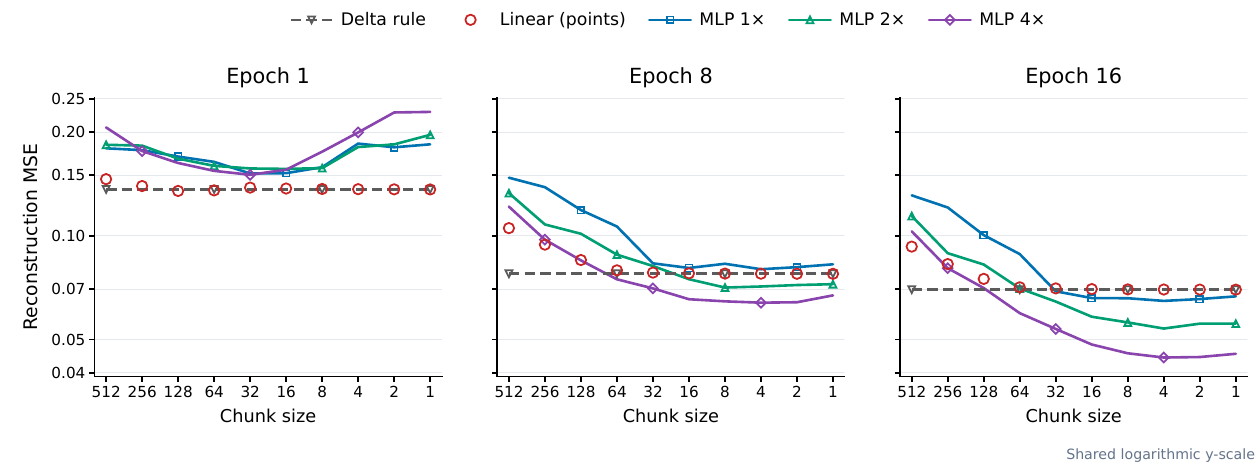}
\caption{Fitting fixed Full-Attention K/V pairs at different replay budgets.}

\label{fig:motivation-one-epoch}
\end{figure}

\noindent\textbf{Nonlinear memories can struggle to fit well within a single pass.}
Greater expressivity does not necessarily translate into better fitting under a limited optimization budget. To examine this, we conduct a toy experiment in which a linear map and a two-layer MLP fit the same fixed K/V sequence extracted from a full-attention model. We compare their reconstruction errors across epochs. As shown in Figure~\ref{fig:motivation-one-epoch}, the linear memory can achieve lower reconstruction error after just one epoch, despite the MLP's greater capacity. The MLP attains lower error at small chunk sizes by epoch 16 (Figure~\ref{fig:motivation-one-epoch}), suggesting that its capacity advantage requires more optimization to emerge under this protocol. However, existing methods such as LaCT~\citep{zhang2025lact}, TTT~\citep{sun2024ttt}, and DeltaNet~\citep{yang2024deltanet} typically use a single pass over the context for computational efficiency, as additional epochs of inner learning would incur substantial overhead. These results motivate more effective optimization of nonlinear memories within a single-pass budget. Appendix~\ref{app:kv-reconstruction} provides the experimental details.


\noindent\textbf{Smaller chunks can improve fitting but incur high latency in serial nonlinear TTT.}
Figure~\ref{fig:motivation-one-epoch} highlights a second constraint: at a fixed replay budget, smaller chunks allow more update steps and can yield lower reconstruction error, although the trend is not strictly monotonic; for the tuned MLP, the benefit is more apparent after multiple passes. A related tradeoff was observed by \citet{sun2024ttt}, who found
that smaller TTT mini-batches reduce language-model perplexity
at the cost of higher latency. In serial nonlinear TTT, each chunk's gradient depends on the memory updated by preceding chunks, so smaller chunks
require more sequential update steps while offering less
parallelism within each step. This computational tradeoff motivates LaCT's use of large chunks \citep{zhang2025lact}. Linear recurrent memories with delta updates offer a way around this tradeoff. Given the keys and targets, each delta update is affine in the preceding memory state, enabling efficient chunkwise parallel evaluation while preserving the exact sequence of tokenwise corrections \citep{yang2024deltanet}. The logical update granularity can therefore remain $C=1$ even when the hardware kernel processes many tokens together.

\section{DeltaTTT: Layerwise Memory Optimization}
\label{sec:method}
Given the difficulty of optimizing a nonlinear multilayer memory as a whole within a single epoch, can we optimize it more effectively layer by layer? This motivates DeltaTTT, which retains a nonlinear multilayer memory while equipping each linear layer with a local target and its own delta update. Given its inputs and targets, each layer admits efficient chunkwise parallel evaluation, while the nonlinear composition and dependencies between layers are preserved. This design aims to combine the expressive capacity of nonlinear memories with effective single-pass optimization and the computational efficiency of the delta rule.

\subsection{Layerwise memory optimization}
\label{sec:targets}

We now introduce DeltaTTT, a layerwise approach to optimizing nonlinear
recurrent memory. For one memory head, let
$k_t,q_t\in\mathbb{R}^{d_k}$ and $v_t\in\mathbb{R}^{d_v}$ denote the
current key, query, and value.
The memory $W$ consists of two dynamic matrices
$A\in\mathbb{R}^{d_v\times d_k}$ and
$B\in\mathbb{R}^{d_v\times d_v}$, connected by
$\phi=\mathrm{SiLU}$. Classical TTT jointly optimizes this two-layer
memory to reconstruct values from keys:
\begin{equation}
f_{A,B}(x)=B\phi(Ax),
\qquad
\ell_t(A,B)=\frac12\|B\phi(Ak_t)-v_t\|_2^2.
\label{eq:joint-loss}
\end{equation}
Our goal is to make this reconstruction task easier to optimize within a single pass. To this end, we decompose it into two local learning problems, one for each layer. The first layer learns to predict $v_t$ directly from $k_t$.
After updating $A$, its prediction is passed through the nonlinearity
to form an intermediate representation $h_t=\phi(A_tk_t)$.
The second layer then learns to reconstruct the same value $v_t$
from $h_t$. In this way, the first layer makes an initial prediction,
and the second provides another learned reconstruction stage that
can compensate for imperfections in the first. Their local objectives are
\begin{equation}
\ell_{A,t}(A)=\frac12\|Ak_t-v_t\|_2^2,
\qquad
\ell_{B,t}(B)=\frac12\|Bh_t-v_t\|_2^2.
\label{eq:local-losses}
\end{equation}
Each objective is linear regression in its own matrix, so a gradient
step yields a delta-rule update. With learned write rates
$\alpha_t$ and $\beta_t$, the complete update and readout are
\begin{align}
A_t
&= A_{t-1}
 + \alpha_t(v_t-A_{t-1}k_t)k_t^\top,
\label{eq:post-a}\\
h_t
&= \phi(A_tk_t),
\label{eq:post-hidden}\\
B_t
&= B_{t-1}
 + \beta_t(v_t-B_{t-1}h_t)h_t^\top,
\label{eq:post-b}\\
o_t
&= B_t\phi(A_tq_t).
\label{eq:readout}
\end{align}
Thus, the query still reads from a nonlinear memory.
Meanwhile, each linear layer is updated using the delta rule,
allowing DeltaTTT to reuse DeltaNet's chunkwise scan algorithm
to compute tokenwise updates in parallel within each chunk
\citep{yang2024deltanet}.
Through layerwise optimization, DeltaTTT combines tokenwise updates
with efficient parallel computation for nonlinear memory, while
potentially benefiting from the easier optimization of individual
linear layers compared with the nonlinear network.

We call formulation above \emph{DeltaTTT-post}, since $h_t$ uses the
updated first-layer state. To avoid waiting for the current $A$ update,
\emph{DeltaTTT-pre} instead uses $h_t=\phi(A_{t-1}k_t)$, leaving the
remaining equations unchanged. Once this representation is available,
the two current-token writes can proceed concurrently. Our ablations
show comparable performance between Pre and Post, making \emph{DeltaTTT-pre} attractive
for its greater potential concurrency. We therefore adopt \emph{DeltaTTT-pre}, as illustrated in
Figure~\ref{fig:DeltaTTT}(b), as our default method.

To clarify how this update and readout scheme relates to existing
methods, Table~\ref{tab:update-schedules} compares the states used
to evaluate updates, the granularity of memory evolution,
and the states used for query readout.
Mini-batch TTT and E$^2$-TTT retain tokenwise state recurrences,
but evaluate gradients at fixed batch- or chunk-start states.
LaCT and our fixed-base parallel TTT baseline instead evolve
memory at chunk boundaries and use chunk-start states for readout.
DeltaNet supports state-dependent tokenwise updates and
current-token readout with a linear memory.
DeltaTTT combines these properties with a nonlinear memory,
using layerwise delta-rule updates to enable chunkwise
parallel computation.
Pseudocode for the chunkwise computation is provided in
Appendix~\ref{app:chunkwise}.

\begin{table*}[htbp]
\centering
\caption{Comparison of memory update schedules.
Gradient/write evaluated at denotes the fast-weight state used
to compute each update; state recurrence denotes the granularity
of state evolution; output computed from denotes the state used
for query readout.
TTT rows assume nonlinear inner models.
Parallel TTT adapts fixed-initialization batch-gradient TTT to our
chunkwise, read-before-write baseline (Eq.~\ref{eq:parallel}).
}
\label{tab:update-schedules}
\small
\setlength{\tabcolsep}{5pt}
\renewcommand{\arraystretch}{1.15}
\begin{tabular}{@{}llllcc@{}}
\toprule
Method
& \shortstack[l]{Gradient/write\\evaluated at}
& \shortstack[l]{State\\recurrence}
& \shortstack[l]{Output\\computed from}
& Nonlinear
& \shortstack{Efficient\\training} \\
\midrule
Token-wise TTT~\citep{sun2024ttt}
& Previous token & Token-wise & Current token
& $\checkmark$ & $\times$ \\

Mini-batch TTT~\citep{sun2024ttt}
& Batch start & Token-wise & Current token
& $\checkmark$ & $\times$ \\

Chunk-TTT ~\citep{zhang2025lact}
& Chunk start & Chunk-wise & Chunk start
& $\checkmark$ & $\checkmark$ \\

E$^2$-TTT~\citep{zhong2026rethinking}
& Chunk start & Token-wise & Chunk start
& $\checkmark$ & $\checkmark$ \\

Parallel TTT
& Initial state & Chunk-wise & Chunk start
& $\checkmark$ & $\checkmark$ \\

DeltaNet~\citep{yang2024deltanet}
& Previous token & Token-wise & Current token
& $\times$ & $\checkmark$ \\

\textbf{DeltaTTT (ours)}
& Previous token & Token-wise & Current token
& $\checkmark$ & $\checkmark$ \\
\bottomrule
\end{tabular}
\end{table*}

\section{Experiments}
\label{sec:experiments}

\subsection{Experimental setup}
\label{sec:setup}

We train on 92.5B tokens from the pretraining split of
Long-Data-Collections. Our DeltaTTT models contain approximately
0.54B parameters, with 24 layers and a hidden dimension of 1024.
Training uses 32K-token sequences and AdamW with a peak learning
rate of $3\times10^{-4}$ and cosine decay.
We implement DeltaTTT on DeltaNet \citep{yang2024deltanet} and
LaCT \citep{zhang2025lact} backbones by replacing their memory modules with our DeltaTTT module.
Baselines include DeltaNet and a variant with a learnable
initial memory $W_0$ ;
Gated DeltaNet \citep{yang2025gated};
serial LaCT and our fixed-base parallel LaCT variant;
and Delta MLP \citep{irie2021going}, which uses layer-specific
keys and values generated from the same input token to update
a multilayer fast-weight memory via the delta rule.
For a fair comparison, all recurrent models use sliding-window
attention (SWA) with a 512-token window.
Following LaCT's practice, we set both the update chunk size
and SWA window size to 512 tokens for all serial and parallel
LaCT baselines.
A full-attention Transformer \citep{vaswani2023attentionneed} serves
as an additional reference.
We evaluate single-needle retrieval using RULER
\citep{hsieh2024ruler} over 4K--32K contexts.
Detailed experimental settings are provided in
Appendix~\ref{app:protocol}.

\subsection{Main results}
\label{sec:main-results}

Table~\ref{tab:main} shows that \method{} improves both training loss and single-needle retrieval across the two backbones. On the LaCT backbone, fixed-base parallel updates outperform serial updates, but \method{} surpasses both while retaining state-dependent correction, including a 12.60 point gain in single-needle retrieval over parallel LaCT. This suggests that a more effective optimization strategy can help realize the potential of nonlinear, state-dependent memory. Improvements over DeltaNet and Gated DeltaNet on the DeltaNet backbone further support the applicability of layerwise optimization across different backbone designs.

\begin{table*}[t]
\caption{Results grouped by backbone at the 92.5B-token budget.
All recurrent configurations use SWA with window 512.
Single denotes the mean over S1/S2/S3.
}
\label{tab:main}
\centering
\small
\setlength{\tabcolsep}{3pt}
\renewcommand{\arraystretch}{1.15}
\resizebox{\textwidth}{!}{%
\begin{tabular}{@{}lrrr*{15}{r}@{}}
\toprule
Model & Params (M) & Train loss $\downarrow$ & Single $\uparrow$
& \multicolumn{5}{c}{S1 $\uparrow$}
& \multicolumn{5}{c}{S2 $\uparrow$}
& \multicolumn{5}{c}{S3 $\uparrow$} \\
\cmidrule(lr){5-9}
\cmidrule(lr){10-14}
\cmidrule(l){15-19}
& & & &
4K & 8K & 16K & 32K & Avg. &
4K & 8K & 16K & 32K & Avg. &
4K & 8K & 16K & 32K & Avg. \\
\midrule
Full attention
& 536.95 & 1.918769 & 72.42
& 100.0 & 100.0 & 99.2 & 95.2 & 98.60
& 95.4 & 91.2 & 95.6 & 88.2 & 92.60
& 50.0 & 27.8 & 22.2 & 4.2 & 26.05 \\
\midrule
\multicolumn{19}{@{}l}{DeltaNet backbone} \\
DeltaNet
& 537.64 & 1.955473 & 44.38
& 99.8 & 98.4 & 97.2 & 91.8 & 96.80
& 83.0 & 32.6 & 10.8 & 2.4 & 32.20
& 7.6 & 4.2 & 2.2 & 2.6 & 4.15 \\
Gated DeltaNet
& 562.80 & 1.967150 & 47.60
& 100.0 & 100.0 & 100.0 & 98.8 & 99.70
& 76.0 & 43.6 & 19.4 & 5.0 & 36.00
& 17.6 & 7.2 & 2.2 & 1.4 & 7.10 \\
\rowcolor{gray!12}
DeltaTTT
& 541.17 & \textbf{1.950079} & \textbf{48.78}
& 100.0 & 100.0 & 100.0 & 99.8 & \textbf{99.95}
& 93.4 & 44.0 & 14.2 & 4.6 & \textbf{39.05}
& 21.4 & 4.6 & 1.6 & 1.8 & \textbf{7.35} \\
\midrule
\multicolumn{19}{@{}l}{LaCT backbone} \\
DeltaNet
& 538.13 & 1.970605 & 25.00
& 85.2 & 64.2 & 34.0 & 4.0 & 46.85
& 50.0 & 29.2 & 10.2 & 3.4 & 23.20
& 11.2 & 4.2 & 2.6 & 1.8 & 4.95 \\
DeltaNet (learned init)
& 539.70 & 1.969120 & 40.52
& 100.0 & 98.8 & 72.0 & 21.4 & 73.05
& 84.6 & 54.6 & 22.2 & 9.6 & 42.75
& 15.0 & 5.2 & 1.6 & 1.2 & 5.75 \\
Delta MLP
& 592.10 & 1.968994 & 4.88
& 16.6 & 5.4 & 3.2 & 1.2 & 6.60
& 13.8 & 6.2 & 1.6 & 1.8 & 5.85
& 4.0 & 2.2 & 1.2 & 1.4 & 2.20 \\
LaCT Serial SwishGLU
& 546.98 & 1.989887 & 28.25
& 56.4 & 52.2 & 41.4 & 29.4 & 44.85
& 79.4 & 21.8 & 9.0 & 2.2 & 28.10
& 32.6 & 9.0 & 3.8 & 1.8 & \textbf{11.80} \\
LaCT Parallel SwishGLU
& 546.98 & 1.984051 & 37.87
& 89.0 & 94.6 & 91.0 & 88.8 & 90.85
& 41.8 & 12.0 & 7.6 & 1.8 & 15.80
& 16.4 & 7.2 & 2.2 & 2.0 & 6.95 \\
\rowcolor{gray!12}
DeltaTTT
& 541.67 & \textbf{1.952344} & \textbf{50.47}
& 99.8 & 99.6 & 97.6 & 69.8 & \textbf{91.70}
& 86.6 & 73.0 & 34.8 & 9.6 & \textbf{51.00}
& 22.6 & 7.8 & 2.6 & 1.8 & 8.70 \\
\bottomrule
\end{tabular}%
}
\end{table*}


\subsection{Context use and position-dependent loss}
\label{sec:context}

\begin{figure}[t]
  \centering
  \includegraphics[width=0.9\linewidth]{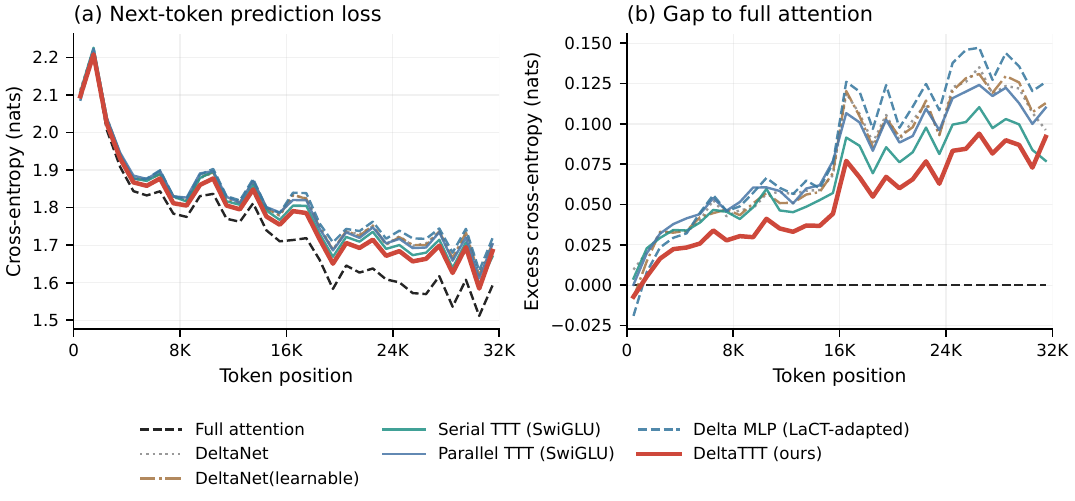}
  \caption{Average Position-wise loss on 43 held-out documents (19 books, 24 papers), each at least 32K tokens long. }
  \label{fig:context-position-loss}
\end{figure}

\paragraph{Position-dependent prediction.} 
We select 19 books and 24 arXiv papers from the Long-Data-Collections, which are excluded from training. Each is long enough to provide 32K prediction targets from a single document prefix. 
We compute teacher-forced next-token loss for each document, 
then average losses across documents in 1K-token bins. 
Figure~\ref{fig:context-position-loss} shows both absolute loss 
and excess loss relative to full attention. 
\method{} has the lowest average loss among the evaluated 
LaCT-backbone recurrent models and reduces the gap to full 
attention, with a larger advantage over parallel LaCT 
at later positions. Full attention retains the lowest overall loss.

\subsection{Training efficiency}
\label{sec:efficiency}

We measure throughput and peak reserved
memory for optimized 0.5B models at a sequence
length of 32K at training stage.
As shown in Table~\ref{tab:efficiency}, \method{} is $37\%$
faster than full attention and uses $31\%$ less peak memory.
It further outperforms serial and parallel LaCT in throughput
by $65\%$ and $28\%$, respectively, with comparable or lower
peak memory usage.

\begin{table}[t]
\caption{Training efficiency relative to full attention ($1.00\times$).
}
\label{tab:efficiency}
\centering
\small
\begin{tabular}{lrr}
\toprule
Model & Throughput/GPU $\uparrow$ & Peak memory/GPU $\downarrow$ \\
\midrule
Full attention & $1.00\times$ & $1.00\times$ \\
DeltaNet & $1.42\times$ & $0.53\times$ \\
LaCT Serial (SwiGLU) & $0.83\times$ & $0.68\times$ \\
LaCT Parallel (SwiGLU) & $1.07\times$ & $0.86\times$ \\
\textbf{\method{}(ours)} & $1.37\times$ & $0.69\times$ \\
\bottomrule
\end{tabular}
\end{table}

\subsection{Ablations}
\label{sec:ablations}

\paragraph{Pre versus Post updates.}
Table~\ref{tab:ablation-structure}(a) compares the Pre and Post
variants on the DeltaNet backbone.
Pre and Post achieve similar training losses and comparable
single-needle retrieval accuracy.
Pre constructs the second-layer write address from the
pre-update state $A_{t-1}$, removing its dependence on the
current first-layer update and enabling concurrent updates
of the two memory modules during inference.

\paragraph{Memory components.}
Starting from the selected DeltaTTT+SWA model in
Table~\ref{tab:main}, we progressively remove memory components
in Table~\ref{tab:ablation-structure}(b).
We first disable online updates to $B$, retaining its learned
initial state $B_0$ and the nonlinear readout.
We then remove the remaining static second layer and the
intermediate nonlinearity, yielding DeltaNet+SWA with a
learnable initial memory $A_0$.
Finally, fixing $A_0=0$ gives the zero-initialized DeltaNet+SWA
baseline, while preserving its online delta-rule updates.
The full model achieves the lowest training loss and highest
retrieval accuracy.
Disabling second-layer updates substantially reduces retrieval
accuracy, and retaining a static second layer performs worse
than the single-layer model.
Removing the learnable initialization further degrades
single-layer retrieval.
These results highlight the benefits of online adaptation
in the second layer and a learned initial memory.

\begin{table}[t]
\caption{Ablations at 0.5B model and 92.5B training tokens,
Removals in (b) are cumulative; 
Single is the mean single-needle retrieval accuracy across S1/S2/S3.}
\label{tab:ablation-structure}
\centering
\small
\setlength{\tabcolsep}{4pt}
\renewcommand{\arraystretch}{1.12}
\begin{tabularx}{\textwidth}{@{}p{0.31\textwidth}Xrr@{}}
\toprule
Model & Configuration & Train loss $\downarrow$ & Single $\uparrow$ \\
\midrule
\multicolumn{4}{@{}l}{\textbf{(a) Write address}} \\
DeltaTTT-Pre (selected)
& Address from $A_{t-1}$ & 1.962162 & 54.32 \\
DeltaTTT-Post
& Address from $A_t$ & 1.960538 & 52.22 \\
\midrule
\multicolumn{4}{@{}l}{\textbf{(b) Progressive component removal}} \\
DeltaTTT (selected)
& Full model
& 1.952344 & 50.47 \\
DeltaTTT
& $-$ $B$ online update
& 1.967580 & 36.02 \\
DeltaNet
& $-$ Learnable $B_0$
& 1.969120 & 40.52 \\
DeltaNet
& $-$ Learnable $A_0$
& 1.970605 & 25.00 \\
\bottomrule
\end{tabularx}
\end{table}

\subsection{Mechanism analysis}
\label{sec:mechanism}

\begin{figure}[t]
\centering
\includegraphics[width=0.6\linewidth]{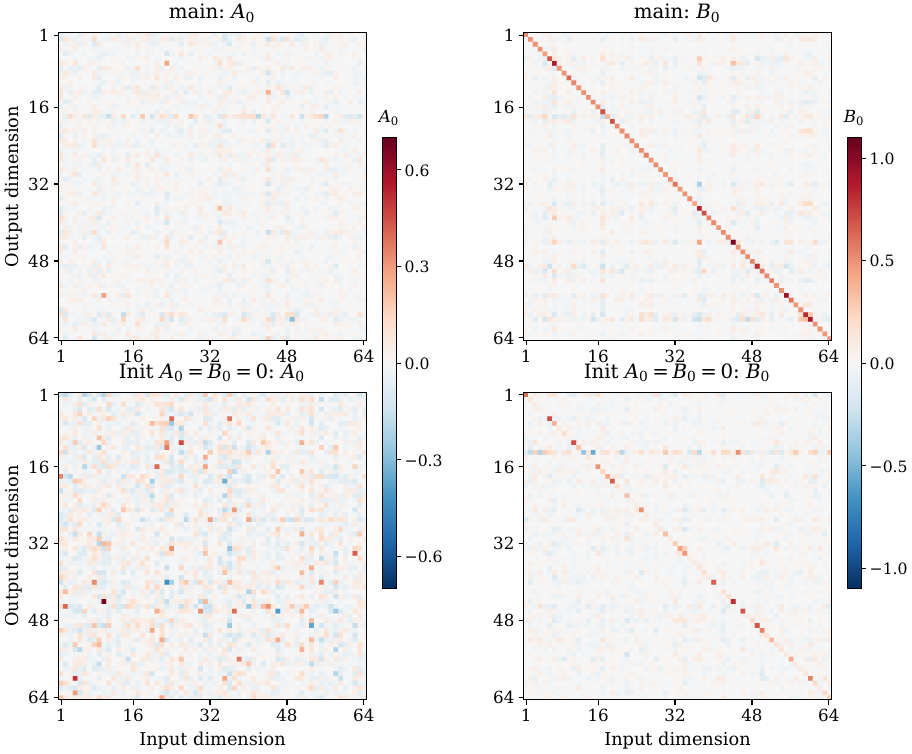}
\caption{Learned $A_0$ and $B_0$ at layer 13,
head 1. Top: main configuration; bottom: the variant with
both matrices initialized to zero.}
\label{fig:native-learned-ab}
\end{figure}

\paragraph{What do the two layers learn?}
To examine what the two layers learn, we visualize the
initial memory matrices $A_0$ and $B_0$ after outer training
(Figure~\ref{fig:native-learned-ab}).
Since our main configuration includes a fixed $0.5I$ offset
in $B_0$, we also inspect a variant in which both $A_0$ and
$B_0$ are initialized to zero.
The learned $B_0$ in this variant also exhibits a pronounced
diagonal structure, suggesting that outer training learns
a prior for refinement.
This is consistent with the intended division of labor:
$A$ produces a value-aligned intermediate representation,
and $B$ further adjusts it to recover the value.

\paragraph{How much do online writes reduce reconstruction loss?}
Table~\ref{tab:kv-reconstruction} compares the initial and final
full-sequence reconstruction errors of DeltaTTT and two DeltaNet
baselines on the same 32K validation sequence.
Each checkpoint is evaluated using its self-generated K/V pairs.
DeltaTTT reduces MSE by 81.88\%, compared with 70.88\% for
zero-initialized DeltaNet and 76.44\% for its
learned-initialization variant.
These within-model reductions show that our layerwise online
updates substantially improve memory reconstruction.

\begin{table}[t]
\caption{Initial and final full-sequence reconstruction errors
on the same 32K validation sequence.
Each checkpoint uses its self-generated K/V features.
Reduction denotes the relative decrease in MSE.}
\label{tab:kv-reconstruction}
\centering
\small
\setlength{\tabcolsep}{4pt}
\begin{tabularx}{\linewidth}{@{}Xrrrr@{}}
\toprule
Model
& \shortstack{Train tokens}
& \shortstack{Initial MSE}
& \shortstack{Final MSE}
& Reduction \\
\midrule
\shortstack[l]{DeltaNet}
& 92.5B & 0.00408555 & 0.00118968 & 70.88\% \\
\shortstack[l]{DeltaNet(learned init)}
& 92.5B & 0.00466836 & 0.00109973 & 76.44\% \\
\shortstack[l]{DeltaTTT}
& 92.5B & 0.00468315 & 0.00084878 & 81.88\% \\
\bottomrule
\end{tabularx}
\end{table}

\section{Related Work}
\label{sec:related}

\paragraph{TTT and nonlinear fast weights.}
TTT was introduced for adapting models to distribution shifts
through self-supervised updates at test time \citep{sun2020ttt}.
For sequence modeling, TTT layers represent recurrent memory
as an inner learner updated through local self-supervised
objectives \citep{sun2024ttt}.
LaCT scales nonlinear fast weights using large update chunks
\citep{zhang2025lact}, while Titans incorporates momentum and
forgetting into online neural memory \citep{behrouz2025titans}.
TNT addresses the chunk-size tradeoff through hierarchical
memories and subsequent fine-tuning at smaller chunk sizes
\citep{li2025tnt}.
E$^2$-TTT preserves token-dependent learning rates, momentum,
and decay in a parallel chunk update under shared chunk-start
gradient evaluation \citep{zhong2026rethinking}.
Modular TTT systematically studies inner-learner design and
finds that deeper fast-weight networks can underperform
shallow alternatives in its evaluated one-step setting
\citep{tang2026modular}.
Complementing this observation, \citet{liu2026kvbinding}
show that better inner KV fitting need not improve downstream
performance and analyze connections between KV-binding TTT
and learned linear attention.
Other work changes the adaptation objective: TTT-E2E uses
next-token prediction and meta-learns an initialization for
adaptation \citep{tandon2025endtoend}, while TTCD supervises
fast weights using a long-window teacher and a short-window
student \citep{context2026distillation}.
TTCD's off-policy implementation constructs updates from
base-weight activations and accumulates them causally,
providing a related instance of fixed-base updating.
In vision, ViT$^3$ studies inner architectures and training
rules \citep{han2026vit3}, while T$^5$ converts pretrained
vision Transformers into TTT models through architectural
and representational alignment \citep{li2026linearizing}.
Our work focuses on layerwise optimization of nonlinear
recurrent memory within a single pass, with reconstruction
error serving as a diagnostic alongside downstream evaluation.

\paragraph{Delta-rule memories and online optimization.}
Linear attention admits a recurrent formulation that
accumulates key--value associations
\citep{katharopoulos2020linear}.
\citet{schlag2021linear} connect linear attention to fast weight
programmers and introduce delta-rule writes that correct
existing associations.
\citet{yang2024deltanet} develop efficient chunkwise evaluation
of this recurrence, and Gated DeltaNet combines delta-rule
corrections with adaptive forgetting \citep{yang2025gated}.
Beyond linear memories, \citet{irie2021going} explore recurrent and multilayer fast
networks, including Delta MLPs that use layer-specific keys
and values generated from the same input token to update
a multilayer fast-weight memory via the delta rule.
Test-time regression and MIRAS provide broader frameworks
for understanding associative memories through their function
classes, objectives, and learning rules
\citep{wang2025regression,behrouz2025miras}.
ATLAS optimizes memory over a recent context window
\citep{behrouz2025atlas}, while MesaNet uses a conjugate-gradient
solver for in-context regression \citep{vonoswald2025mesanet}.
DeltaTTT assigns explicit value-reconstruction targets to
each linear layer and constructs the second-layer write
address from the first layer's prediction.
Its local objectives replace joint inner-loop optimization,
while the nonlinear intermediate representation couples the
two memory stages and each stage reuses chunkwise delta-rule
evaluation.

\paragraph{Layerwise and local learning.}
Difficulties in training deep networks motivated early methods
that learned representations one layer at a time.
Deep belief networks and greedy layerwise pretraining used
successive shallow learning problems to initialize deep models
before global fine-tuning
\citep{hinton2006fast,bengio2007greedy}.
Stacked denoising autoencoders similarly learned deep
representations through local reconstruction objectives
\citep{vincent2010stacked}.
Later work explored layerwise targets
\citep{lee2015difference}, synthetic gradients that relax
update dependencies \citep{jaderberg2017dni}, and greedy
supervised learning at ImageNet scale
\citep{belilovsky2019greedy}.
Decoupled Greedy Learning further uses local objectives to
relax update locking between network modules
\citep{belilovsky2020decoupled}.
Our motivation shares the idea of making a deep composition
easier to optimize through simpler local learning problems.
In DeltaTTT, these problems govern the online updates of
context-dependent memory states, while the outer language
model remains trained end to end.

\section{Conclusion}
\label{sec:conclusion}

Our findings suggest that greater memory expressivity alone does not guarantee effective learning within a single pass. We introduced \method{}, which decomposes nonlinear memory optimization into local linear reconstruction objectives with layerwise delta-rule updates, preserving nonlinear composition and enabling chunkwise parallel evaluation. Experiments on two backbones demonstrate improvements in language modeling and associative retrieval. 

\section{Limitations}
\label{sec:limitations}

Our current DeltaTTT remains slower and uses more memory than
DeltaNet. Since the kernels for our two-layer memory are not yet fully
optimized, we believe there is room to reduce this overhead through
further implementation improvements. In addition, our experiments
only consider two-layer MLP memories with a hidden expansion ratio
of $1\times$. Extending the approach to wider intermediate
representations and deeper nonlinear memories remains an important
direction for assessing whether layerwise optimization can effectively
exploit greater memory capacity. Finally, our evaluation is limited
to relatively small models and a bounded training-token budget.
Further experiments with larger models and more training data are
needed to establish the scaling potential of \method{}.

\subsection*{AI use statement}
Generative AI assisted with organizing the manuscript and drafting text and
equations from existing experiment records in this revision.
No new training or evaluation was performed for this manuscript edit.

\subsection*{Reproducibility statement}
Section~\ref{sec:method} specifies the memory updates and readout,
and Section~\ref{sec:experiments} describes the experimental setup
and evaluation.
Appendix~\ref{app:chunkwise} provides the chunkwise algorithms
for DeltaNet and DeltaTTT-pre.
Appendix~\ref{app:kv-reconstruction} details the fixed-KV reconstruction
experiments, including data selection, memory architectures,
learning-rate tuning, and evaluation.
Appendix~\ref{app:protocol} documents the shared training configuration,
backbone-specific implementations, and main-table DeltaTTT settings.
Appendix~\ref{app:initialization} reports ablations of the learned
initial memory states.

\bibliographystyle{iclr2027_conference}
\bibliography{iclr2027_conference}

\clearpage
\appendix

\section{Chunkwise Scan of DeltaTTT}
\label{app:chunkwise}

Algorithms~\ref{alg:chunkwise-deltanet} and
\ref{alg:chunkwise-deltattt} describe one memory head.
The sequence is divided into $N$ chunks of up to $C$ tokens,
with $Q_c,K_c,V_c$ stacking token vectors as rows.
$W$ denotes DeltaNet's memory, while $A$ and $B$ denote
DeltaTTT's first- and second-layer memories, respectively.
$\boldsymbol{\eta}_c$, $\boldsymbol{\alpha}_c$, and
$\boldsymbol{\beta}_c$ contain the tokenwise write rates
for $W$, $A$, and $B$, respectively.
For an incoming memory $M$, the chunkwise writes are
\begin{equation}
\operatorname{Delta}(K,V,\boldsymbol{r};M)
=
\bigl[I+D\operatorname{tril}(KK^\top,-1)\bigr]^{-1}
D(V-KM^\top),
\qquad D=\operatorname{diag}(\boldsymbol{r}).
\label{eq:delta-chunk}
\end{equation}
Here $I$ is the identity of the chunk length, and the inverse
denotes a triangular solve.
$\operatorname{tril}(X)$ includes the diagonal, whereas
$\operatorname{tril}(X,-1)$ excludes it.
$\phi=\silu$ is applied elementwise.
In DeltaTTT-pre, $H_c$ and $R_c$ are the first-layer
pre-update key and post-update query representations,
respectively.
$U_c$ stores write-rate-scaled residuals, and
$[O_1;\ldots;O_N]$ concatenates the output chunks.

\begin{algorithm}[htbp]
\caption{Chunkwise DeltaNet: one scan}
\label{alg:chunkwise-deltanet}
\small
\begin{algorithmic}[1]
\Require Chunks $\{Q_c,K_c,V_c,\boldsymbol{\eta}_c\}_{c=1}^N$;
         initial memory $W_0$
\State $W\gets W_0$
\For{$c=1,\ldots,N$}
  \State $U_c\gets\operatorname{Delta}(K_c,V_c,\boldsymbol{\eta}_c;W)$
  \State $O_c\gets Q_cW^\top+\operatorname{tril}(Q_cK_c^\top)U_c$
  \State $W\gets W+U_c^\top K_c$
\EndFor
\State \Return $[O_1;\ldots;O_N],W$
\end{algorithmic}
\end{algorithm}

\begin{algorithm}[htbp]
\caption{Chunkwise DeltaTTT-pre: two successive scans}
\label{alg:chunkwise-deltattt}
\small
\begin{algorithmic}[1]
\Require Chunks $\{Q_c,K_c,V_c,\boldsymbol{\alpha}_c,
         \boldsymbol{\beta}_c\}_{c=1}^N$;
         initial memories $A_0,B_0$
\State $A\gets A_0$; $B\gets B_0$
\For{$c=1,\ldots,N$} \Comment{First scan: one update, two reads}
  \State $U_c\gets\operatorname{Delta}(K_c,V_c,\boldsymbol{\alpha}_c;A)$
  \State $H_c\gets\phi\!\left(
    K_cA^\top+\operatorname{tril}(K_cK_c^\top,-1)U_c\right)$
  \State $R_c\gets\phi\!\left(
    Q_cA^\top+\operatorname{tril}(Q_cK_c^\top)U_c\right)$
  \State $A\gets A+U_c^\top K_c$
\EndFor
\For{$c=1,\ldots,N$} \Comment{Second scan: one update, one read}
  \State $U_c\gets\operatorname{Delta}(H_c,V_c,\boldsymbol{\beta}_c;B)$
  \State $O_c\gets R_cB^\top+\operatorname{tril}(R_cH_c^\top)U_c$
  \State $B\gets B+U_c^\top H_c$
\EndFor
\State \Return $[O_1;\ldots;O_N],A,B$
\end{algorithmic}
\end{algorithm}

\section{Fixed-KV Reconstruction Experiments}
\label{app:kv-reconstruction}
\label{app:standalone-memory}

Figure~\ref{fig:motivation-one-epoch} compares linear and nonlinear
memories fitted to fixed K/V caches from a frozen 0.5B
full-attention model trained on approximately 10B tokens.
We use eight 32K sequences and all 24 layers, concatenating
heads to a feature dimension of 1,024.
Four sequences are used to select learning rates, and the
remaining four are used to report results.

We compare a zero-initialized linear memory with a two-layer
SiLU MLP of hidden width 1x/2x/4x and Xavier-uniform initialization.
Both memories are optimized using ordered SGD replay,
with chunk sizes ranging from 1 to 512 in powers of two.
Each chunk $\mathcal C$ of size $C$ produces one update under
the key-normalized squared loss:
\begin{equation}
\mathcal L_{\mathcal C}(\theta)
=
\frac{1}{2C}
\sum_{i\in\mathcal C}
\frac{\|f_\theta(k_i)-v_i\|_2^2}
{\|k_i\|_2^2}.
\label{eq:motivation-normalized-loss}
\end{equation}
The MLP updates both matrices simultaneously, and initializations
are shared across chunk sizes.

Learning rates are tuned separately for each model and chunk
size using reconstruction MSE after 16 epochs; the MLP uses
a separate rate for each layer.
The selected rates remain fixed for evaluation.
We report MSE reconstruction loss over each complete fitted
sequence, averaged across the four reporting sequences and all layers.
The tokenwise Delta reference corresponds to the linear memory
with $C=1$.
Equal numbers of passes do not imply equal computational cost:
smaller chunks require more updates.

\section{Main-Table Experimental Details}
\label{app:protocol}

\paragraph{Shared training setup.}
Table~\ref{tab:training-config} summarizes the shared training
configuration. All recurrent models use SWA with a 512-token window.

\begin{table}[htbp]
\caption{Shared training configuration}
\label{tab:training-config}
\centering
\small
\setlength{\tabcolsep}{6pt}
\renewcommand{\arraystretch}{1.1}
\begin{tabularx}{\linewidth}{@{}lX@{}}
\toprule
Setting & Value \\
\midrule
Training data & Long-Data-Collections, 92.5B tokens \\
Model dimensions & 24 layers; hidden width 1,024; FFN width 4,096 \\
Vocabulary and embeddings & 131,072 tokens; tied input/output embeddings \\
Sequence length & 32,768 tokens \\
Optimizer & AdamW, $(\beta_1,\beta_2)=(0.9,0.95)$,
$\epsilon=10^{-8}$ \\
Weight decay / gradient clipping & 0.1 / 1.0 \\
Learning rate & Peak $3\times10^{-4}$;
cosine decay to $3\times10^{-5}$ \\
Warmup & Approximately 1.05B tokens \\
\bottomrule
\end{tabularx}
\end{table}

\paragraph{Backbone implementations.}
We implement DeltaTTT within both DeltaNet and LaCT backbones,
replacing their recurrent memory modules while retaining
backbone-specific feature processing and output integration.
The main differences concern positional encoding and memory
output modulation: the DeltaNet backbone uses RoPE only in
the SWA branch and directly adds the RMS-normalized memory
output to the SWA output, whereas the LaCT backbone also
applies RoPE to memory Q/K and uses an input-dependent scale
to modulate the normalized memory output.
LaCT additionally retains learned affine transforms on
the memory inputs.

\paragraph{DeltaTTT configuration.}
We align the main architectural settings with our DeltaNet
baseline: both implementations use 16 memory heads of width 64,
causal depthwise Conv4 with SiLU, L2-normalized memory Q/K.
The memory and SWA branches share Q/K/V projections, and
their outputs are summed before the output projection.
Both main-table DeltaTTT models use Pre updates for memory layers.
The learned initial states use
$A_0\sim\mathcal{N}(0,0.02^2)$ and
$B_0=LR+0.5I$, where $L,R$ are learned rank-32 factors.



\section{Initial-State Ablation}
\label{app:initialization}

Table~\ref{tab:ablation-initialization} shows that different
learnable initializations yield similar training loss and
single-needle retrieval accuracy.

\begin{table}[htbp]
\caption{Initial-state ablations for DeltaTTT on the DeltaNet
backbone. Single denotes average single-needle accuracy (\%).}
\label{tab:ablation-initialization}
\centering
\small
\setlength{\tabcolsep}{4pt}
\renewcommand{\arraystretch}{1.2}
\begin{tabularx}{\textwidth}{@{}Xlrr@{}}
\toprule
Model & Initial states & Train loss $\downarrow$ & Single $\uparrow$ \\
\midrule
DeltaTTT
& $A_0=B_0=0$ (learnable)
& 1.95 & 47.70 \\
DeltaTTT
& $(A_0)_{ij},(B_0)_{ij}\sim\mathcal{N}(0,0.02^2)$
& 1.95 & 48.00 \\
DeltaTTT
& $(A_0)_{ij}\sim\mathcal{N}(0,0.02^2),\ B_0=LR+0.5I$
& 1.95 & 48.78 \\
\bottomrule
\end{tabularx}
\end{table}

\end{document}